\documentclass[11pt,a4paper]{article}

\usepackage{authblk}
\usepackage{relsize}

\usepackage[numbers,sort&compress,longnamesfirst,sectionbib]{natbib}
\usepackage[noend]{algpseudocode}
\usepackage[xspace]{ellipsis}
\usepackage{IEEEtrantools}
\usepackage{algorithm,dsfont}
\usepackage{amsfonts}
\usepackage{balance}
\usepackage{booktabs}                               
\usepackage{color}   
\usepackage{graphicx}
\usepackage{mathtools}
\usepackage{multirow}
\usepackage{nicefrac}
\usepackage{subcaption}
\usepackage{tabularx}
\usepackage{tikz}
\usepackage{url}
\usepackage{xcolor}
\usepackage{xspace}
\usetikzlibrary{arrows,patterns,plotmarks,shapes,snakes,er,3d,automata,backgrounds,topaths,trees,petri,mindmap}

\DeclareSymbolFont{symbolsC}{U}{pxsyc}{m}{n}
\DeclareMathSymbol{\colonequals}{\mathrel}{symbolsC}{"42}

\newcommand{\etal}{et al.\ }

\newcommand{\tmax}{\ensuremath{t_{\max}}\xspace}

\newcommand{\ignore}[1]{}

\newcommand{\M}[1]{\mathbf{#1}}

\newcommand{\R}{\mathbb{R}}
\newcommand{\SPD}[1][p]{\mathit{Sym}_+(#1)}

\newcommand{\Sym}[1][p]{\mathit{Sym}(#1)}
\newcommand{\T}{\ensuremath{\mathsf{T}}}

\newcommand{\dist}[1]{\mathrm{dist}_{\mathrm{#1}}}

\newcommand{\frob}[1]{\left \| #1 \right \|_{\mathrm{F}}}

\newcommand{\ve}[1]{\mathbf{#1}}

\let\M\relax 

\newenvironment{changemargin}[2]{%
  \begin{list}{}{%
      \setlength{\topsep}{0pt}%
      \setlength{\leftmargin}{#1}%
      \setlength{\rightmargin}{#2}%
      \setlength{\listparindent}{\parindent}%
      \setlength{\itemindent}{\parindent}%
      \setlength{\parsep}{\parskip}%
    }%
  \item[]}
  {\end{list}}

\newenvironment{keywords}{%
  \begin{changemargin}{\leftmargin}{\leftmargin}
    \small\noindent\emph{Keywords}: }
  {\end{changemargin}}

\begin{document}

\title{Evolutionary Image Composition Using Feature Covariance Matrices}

\author{Aneta Neumann}
\author{Zygmunt L. Szpak}
\author{Wojciech Chojnacki}
\author{Frank Neumann}
\affil{%
  School of Computer Science, The University of Adelaide, SA 5005,
  Australia}
\affil{%
  \texttt{\smaller[0.5]
  \{aneta.neumann, zygmunt.szpak, wojciech.chojnacki, frank.neumann\}@adelaide.edu.au}
}


\date{}

\maketitle

\begin{abstract}
  Evolutionary algorithms have recently been used to create a wide range of artistic work. In this paper, we propose a new approach for the composition of new images from existing ones, that retain some salient features of the original images. We introduce evolutionary algorithms that create new images based on a fitness function that incorporates feature covariance matrices associated with different parts of the images. This approach is very flexible in that it can work with a wide range of features and enables targeting specific regions in the images. For the creation of the new images, we propose a population-based evolutionary algorithm with mutation and crossover operators based on random walks.  Our experimental results reveal a spectrum of aesthetically pleasing images that can be obtained with the aid of our evolutionary process.
\end{abstract}



\section{Introduction}

Evolutionary algorithms have been used in a wide range of areas to come up with novel solutions. This includes the classical task of computing high performing solutions for combinatorial optimisation problems, new designs in the area of engineering, as well as creative solutions in the areas of music and art~\cite{romero08:_art_artif_evolut,antunes15:_writin_readin_artis_comput_ecosy,lambert13,mccormack12:_comput_creat}.  In the research area of evolutionary algorithms and art, the primary aim is to evolve artistic and creative outputs through an evolutionary process~\cite{greenfield15:_avoid,al-rifaie13:_swarm_paint_colour_atten,machado14:_seman,trist11}.
The composition of images has gained some attention in the literature and mainly focuses on using textures. In these works, genetic programming is used to synthesise procedural texture formulas~\cite{ross06:_evolut_image_synth_using_model_aesth}.
Furthermore, genetic programming has been used to evolve functions constructing images based on aesthetic features
~\cite{heijer10:_using,heijer14:_inves}.

This paper presents an approach to carry out image composition directly based on two given images.  Recently, evolutionary processes have been used to create artistic videos through evolutionary image transition. In evolutionary image transition, a source image is transferred into a target image by an evolutionary algorithm, with the images constructed during this process forming frames of a video sequence \cite{neumann16:_evolut_proces_image_trans_conjun,neumann17:_evolut}.  Here, we propose a variant of evolutionary image transition. We use the evolutionary process to create a composition of two images $S$ and $T$. The key element to achieve this goal is to use the concept of a so-called region covariance descriptor, which is well known and well studied in the field of computer vision.  Region covariance descriptors can be used to describe an image based on various sets of features related to different regions of the image. Given two images $S$ and $T$, we aim to evolve an image $X$ which consists of a mixture of pixels from $S$ and $T$ and is ``close'' to both $S$ and $T$ in terms of covariance descriptors for both images.  One of the main contributions of this paper is in finding a way to incorporate region covariance descriptors into an appropriate fitness function.  We investigate the impact of different choices of image features on the aesthetic properties of images created via evolutionary image composition. We also explore the aesthetic impact of weights with which various region covariance descriptors can be combined together to form a fitness function.

The fitness function based on region covariance descriptors is minimised using an evolutionary algorithm which constructs sets of composed images. A key element in our evolutionary algorithm are random walk algorithms used previously in the context of evolutionary image transition. A random walk can be run as part of a mutation phase to obtain transitional images between $S$ and $T$. In the case of crossover, a random walk can be used to generate a crossover mask which is then used to produce an offspring from two parent individuals. The random walk operator introduced in \cite{neumann17:_evolut} depends on a parameter \tmax which determines how many steps a mutation operator performs. Here, with a view to developing an efficient process, we allow for self-adaptation of \tmax during the evolutionary computation. This permits increase of \tmax when the algorithm is making sufficient progress, and decrease of \tmax when the progress is low.  As a result, our evolutionary algorithm adapts to the different stages of the optimisation process. Self-adaption is a key element of evolutionary algorithms for continuous optimisation problems. Recently, Doerr and Doerr~\cite{doerr15:_optim} have modified this notion for the discrete case and shown that a cogent use of the modification significantly speeds up the optimisation for simple benchmark functions.  We use the self-adaptation mechanism in such a fashion that \tmax decreases by a factor when an offspring is rejected, and increases by a factor when an offspring is accepted.

In our experimental investigations we explore how different parameters influence the artistic merit of the resulting images. We explore the role of image features, distances between covariance matrices and region weighting schemes. One of our key innovations is to utilise a measure of visual attention to promote the incorporation of salient regions from both images into the new synthesised images. 

The paper is structured as follows. Section~\ref{sec:RDes} details some prerequisites concerning a covariance descriptor. Section~\ref{sec:fit} presents a fitness function that we propose for image composition. Section~\ref{sec:ea} introduces an evolutionary algorithm including the mutation and crossover operators based on random walks. In Section~\ref{sec:exp}, we present the results of our experimental investigations, and finally finish with a short discussion and some conclusions.

\section{Region Covariance Descriptor}
\label{sec:RDes}

In computer vision various image descriptors have been proposed and used to capture essential characteristics of particular classes of images.  Amongst these the \emph{region covariance descriptor}, introduced by Tuzel \etal \cite{tuzel06:_region_covar}, has proved particularly useful for various computer vision tasks including object tracking~\cite{porikli06:_covar_track_model_updat_based_lie_algeb}, pedestrian detection~\cite{tuzel08:_pedes_rieman}, action recognition~\cite{guo10:_action_recog_using_spars_repres}, and medical imagining~\cite{khan14:_geodes_geomet_mean_region_covar}.  The region covariance descriptor is constructed based on a feature mapping that associates with each pixel in the image a finite-dimensional vector of numerical features such as intensity, colour, gradients, or filter responses.  Once a feature mapping is specified, any region in the image gives rise to the covariance matrix of the feature mapping restricted to this region.  More formally, let $X = (X_{ij})$ be a colour image with a corresponding carrier $\mathit{\Omega} = \{(i,j) \mid \text{$1 \leq i \leq m$ and $1 \leq j \leq n$} \}$ of pixel locations. Each matrix entry $X_{ij}$ is a triplet of integers representing colour information as coordinates in some colour space.  Given a region of interest $\mathcal{R} \subset \mathit{\Omega}$ and a feature mapping $\phi \colon \mathit{\Omega} \to \R^p$, the corresponding \emph{region covariance} matrix is given by
\begin{displaymath}
  \M{\Lambda}_{\mathcal{R}}
  = \frac{1}{|R|-1}
  \sum_{(i,j) \in \mathcal{R}}
  (\phi(i,j)  - \ve{\mu}_{\mathcal{R}})(\phi(i,j)  - \ve{\mu}_{\mathcal{R}})^{\T},
\end{displaymath}
where $\ve{\mu}_{\mathcal{R}} = |\mathcal{R}|^{-1}\sum_{(i,j) \in R}\phi(i,j)$ and $|\mathcal{R}|$ denotes the number of pixels in the region of interest.  The matrix $\M{\Lambda}_{\mathcal{R}}$ describes the variations of the length-$n$ feature vectors $\phi(i,j)$ as $(i,j)$ varies over the region $\mathcal{R}$.  An example feature mapping, used for human detection, is
\begin{align*}
  \phi(i,j)
  & =
    \left[
    \vphantom{
    \sqrt{\big(\tfrac{\partial I}{\partial i}\big)^2_{ij} +
    \big(\tfrac{\partial I}{\partial j}\big)^2_{ij}}
    }
    i, j, I_{ij},
    \big(\tfrac{\partial I}{\partial i}\big)_{ij},
    \big(\tfrac{\partial I}{\partial j}\big)_{ij},
    \big(\tfrac{\partial^2 I}{\partial i^2}\big)_{ij},
    \big(\tfrac{\partial^2 I}{\partial j^2}\big)_{ij},
  \right.
  \\
  & \quad
    \left.
    \sqrt{\big(\tfrac{\partial I}{\partial i}\big)^2_{ij} + \big(\tfrac{\partial I}{\partial j}\big)^2_{ij}},
    \tan^{-1}\left({\big|\big(\tfrac{\partial I}{\partial i}\big)_{ij}\big|}\big/{\big|\big(\tfrac{\partial I}{\partial j}\big)_{ij}\big|}\right)
  \right]^\T,
\end{align*}
where $I_{ij}$ is the image intensity at $(i,j)$, $\big(\frac{\partial I}{\partial i}\big)_{ij}$, $\big(\frac{\partial^2 I}{\partial i^2}\big)_{ij}$, \dots are intensity derivatives at $(i,j)$, and the last two terms are the magnitude of edge response and the edge orientation at $(i,j)$. In line with the convention adopted by MATLAB's \texttt{rgb2gray} function, the image intensity can be suitably defined by
\begin{displaymath}
  I_{ij} = 0.2989 X^R_{ij} + 0.5870 X^G_{ij} + 0.1140 X^B_{ij},
\end{displaymath}
where $X_{ij} = [X^R_{ij}, X^G_{ij}, X^B_{ij}]^\T$ is the decomposition of the colour vector $X_{ij}$ into RGB components.  In this work we explore several candidate feature mappings obtained by selecting component elements from the set of features described in Table~\ref{tab:1}.

\begin{table}[!t]
  \renewcommand{\arraystretch}{1.3}
  \caption{Description of Potential Features for $\phi$}
  \label{tab:1}
  \centering \scalebox{0.8}{
    \begin{tabularx}{\textwidth}{llX}
      \hline
      \multicolumn{2}{c}{\textbf{Notation}} & \textbf{Description}  \\
      \hline
      \multirow{2}{*}{\emph{ij}} & $i$ & vertical spatial coordinate \\
                                            & $j$ & horizontal spatial coordinate \\
      \hline
      \multirow{3}{*}{\emph{rgb}} & $r$ & red channel \\
                                            & $g$ & green channel \\
                                            & $b$ & blue channel \\
      \hline
      \multirow{2}{*}[-1.2em]{$\partial$} & $\big|\frac{\partial I}{\partial i}\big|$ & magnitude of first-order partial derivative
                                                             in horizontal direction \\
                                            & $\big|\frac{\partial I}{\partial j}\big|$ & magnitude of first-order partial derivative
                                                               in vertical direction \\
      \hline
      \multirow{3}{*}[-2.0em]{$\partial^2$} & $\big|\tfrac{\partial^2 I}{\partial i^2}\big|$ & magnitude of second-order partial derivative
                                                                in horizontal direction  \\
                                            & $\big|\tfrac{\partial^2 I}{\partial j^2}\big|$ & magnitude of second-order partial derivative
                                                                in vertical direction \\
                                            & $\big|\tfrac{\partial^2 I}{\partial i \partial j}\big|$ &
                                                                magnitude of second-order mixed partial derivative \\
      \hline
      \multirow{2}*[-0.4em]{\emph{edge}} & $\sqrt{\big(\frac{\partial I}{\partial i}\big)^2 + \big(\frac{\partial I}{\partial j}\big)^2}$
                                 & magnitude of edge response \\
                                            & $\tan^{-1}\left({\big|\frac{\partial I}{\partial i}\big|}\big/{\big|\frac{\partial I}{\partial j}\big|}\right)$ & edge orientation \\
      \hline
      \multirow{3}*{\emph{hsv}} & $h$ & hue (HSV colour space) \\
                                            & $s$ & saturation  (HSV colour space) \\
                                            & $v$ & value  (HSV colour space) \\
      \hline
    \end{tabularx} 
  }
\end{table}

There are several advantages of using covariance matrices as region descriptors.  The feature mapping proposes a natural way of fusing multiple features which might be correlated.  A single covariance matrix extracted from a region is usually enough to match the region in different views and poses.  The noise corrupting individual samples are largely filtered out through averaging which is intrinsic to the process of covariance computation.  The covariance descriptors are low dimensional---each matrix $\M{\Lambda}_{\mathcal{R}}$ has only $p(p+1)/2$ different entries ($p$ is often less than 10), which is a number significantly smaller than the number of histogram bins (going into hundreds) or of raw pixels (going into thousands) used by other descriptors.  Moreover, each $\M{\Lambda}_{\mathcal{R}}$ does not preserve information regarding the ordering and the number of underlying grid points.  This implies a certain degree of scale and rotation invariance over the regions in different images.  However, it should be noted that this near invariance is largely reduced if the feature mapping contains explicit information regarding the orientation of points, such as the gradient of image intensity.  The same is true for illumination.

Covariance matrices are positive-definite and one can speak about a distance between a pair of covariance matrices once a distance measure is defined between members of the set of all real positive-definite matrices.  Let $\Sym$ denote the set of all $p \times p$ symmetric real matrices, and let $\SPD$ denote the subset of $\Sym$ comprised of all $p \times p$ positive-definite matrices in $\Sym$.  $\SPD$ can be endowed with a variety of distance measures \cite{jayasumana13:_kernel_method_rieman_manif_symmet, jayasumana:_kernel_rieman_gauss_rbf}.  In what follows we shall consider three specific distances.  One is the \emph{Euclidean} metric given by
\begin{displaymath}
  \dist{E}(\M{P}, \M{Q}) = \frob{\M{P} - \M{Q}},
\end{displaymath}
where $\frob{\cdot}$ denotes the Frobenius norm, and $\M{P}$ and $\M{Q}$ are members of $\SPD$.  Another is the \emph{Log-Euclidean} metric given by
\begin{displaymath}
  \dist{L}(\M{P}, \M{Q}) = \frob{\log \M{P} - \log \M{Q}},
\end{displaymath}
where $\log$ denotes the principal matrix logarithm \cite{MRM:MRM20965}.  (For an invertible real matrix without eigenvalues on the negative real axis, there always exists a unique real matrix logarithm, called the principal logarithm, whose eigenvalues lie in the strip $\{z \in \mathbb{C} \mid -\pi < \mathrm{Im} z < \pi \}$; cf.\ \cite[Theorem 1.31]{higham08:_funct_matric}.)  Yet another distance measure is the \emph{affine-invariant} metric given by
\begin{displaymath}
  \dist{A}(\M{P}, \M{Q}) = \frob{\log(\M{P}^{-1}\M{Q})}
  = \frob{ \log \big( \M{P}^{-1/2} \M{Q}  \M{P}^{-1/2} \big)}
\end{displaymath}
(cf.\ \cite[Chap.  XII]{lang99:_fundam}).  The label ``affine-invariant'' reflects the fact that $\dist{A}$ is invariant under each mapping of the form $\M{P} \mapsto \M{A} \M{P} \M{A}^\T$, where $\M{A}$ is a real invertible matrix $\M{A}$; that is,
\begin{displaymath}
  \dist{A}(\M{P}, \M{Q}) = \dist{A}(\M{A}  \M{P}  \M{A}^\T, 
  \M{A}  \M{Q}  \M{A}^\T)
\end{displaymath}
for all $\M{P}, \M{Q} \in \SPD$ and all invertible $p \times p$ matrices $\M{A}$.  The affine-invariant metric can alternatively be written as
\begin{equation}
  \label{eq:1}
  \dist{A}(\M{P}, \M{Q}) =
  \left(\sum_{i=1}^p
    \log^2\lambda_i(\M{P}^{-1}\M{Q})
  \right)^{\frac{1}{2}},
\end{equation}
where $\lambda_i(\M{P}^{-1}\M{Q})$, $1 \leq i \leq p$, are the eigenvalues of $\M{P}^{-1}\M{Q}$.  As the matrix $\M{P}^{-1}\M{Q}$ is similar to the symmetric matrix $\M{P}^{-1/2} \M{Q} \M{P}^{-1/2}$, the eigenvalues $\lambda_i(\M{P}^{-1}\M{Q})$ are all positive and hence the right-hand side of \eqref{eq:1} is well defined for all $\M{P}$ and $\M{Q}$ in $\SPD$.

\section{Covariance-based Fitness Function}
\label{sec:fit}
\begin{figure}
  \centering 
  \includegraphics[width=2in]{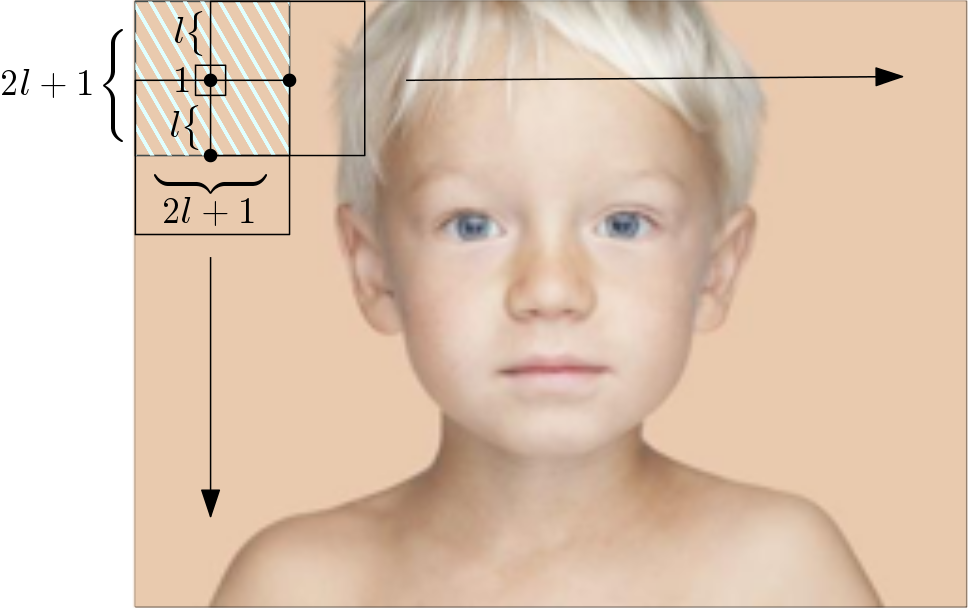}
  \caption{Illustration of overlapping square regions of interest.}
  \label{fig:boy-demo}
\end{figure}

Given two images $S$ and $T$ both of size $m \times n$, we define the fitness of an image $X$ of size $m \times n$ with respect to $S$ and $T$.  The fitness of the mixed image $X$ is evaluated using region covariance descriptors that characterise how similar parts of the generated image are to both input images. The location and shape of the regions associated with the covariance descriptors can be chosen arbitrarily and are based on the personal preference of the artist. In this work we limit ourselves to square regions of interest arranged in a grid-like manner. In particular, we consider regions $\mathcal{R}_{(c,d)} = \{(i,j) \mid |i-c| \le l, |j-d| \le l \}$, which represent squares of size $(2l+1) \times (2l+1)$ centred at $(c,d)$, where $(c,d)$ runs over a grid of pixels $\mathcal{G}$. The grid is defined as
\begin{align*}
  \mathcal{G}  =  \left\{ (c,d)  \left| \;
  \begin{IEEEeqnarraybox}[][c]{l.c.l} 
    \IEEEstrut
    c  & =  &  (l+1) + pl, \; p = 0,1,\ldots, \left \lfloor{\frac{m-l}{l}}\right \rfloor - 1  \\
    d   & = & (l+1) + ql, \; q = 0,1,\ldots, \left \lfloor{\frac{n-l}{l}}\right \rfloor -1  
    \IEEEstrut
  \end{IEEEeqnarraybox}
              \right.
              \right\},
\end{align*}
which results in half-overlapping square regions of interest (see Figure~\ref{fig:boy-demo}).  

To measure the similarity between the generated image and the input images we propose the fitness function
\begin{align*}
  \label{fitnessFunction}
  f(X,S,T) & =
             \sum_{(c,d) \in \mathcal{G}}   \left(  w_{(c,d)}^{S}
             \mathrm{dist}\left(\M{\Lambda}_{\mathcal{R}_{(c,d)}}^{X},
             \M{\Lambda}_{\mathcal{R}_{(c,d)}}^{S} \right) \right.
  \\ 
           & \mathrel{\phantom{=}}
             \left. \mathrel{+} w_{(c,d)}^{T} \mathrm{dist}\left(\M{\Lambda}_{\mathcal{R}_{(c,d)}}^{X},
             \M{\Lambda}_{\mathcal{R}_{(c,d)}}^{T} \right) \right),  
\end{align*}
where $\mathrm{dist}$ is one of the distance functions described in Section \ref{sec:RDes}, and $w_{(c,d)}^{S} \in [0,1]$ and $w_{(c,d)}^{T}\in [0,1]$ are  weights associated with the region $\mathcal{R}_{(c,d)}$ that can be used to emphasise the local contribution of the images $S$ and $T$ in the mixing process.

In addition to the fitness function, we put forward a constraint.  Let $c_S(X) =|\{X_{ij} \mid X_{ij}=S_{ij}\}|$ be the number of pixels in $X$ that are set to $S$ and $c_T(X)=|\{X_{ij} \mid X_{ij}=T_{ij}\}|$ be the number of pixels where $X$ and $T$ agree. We minimise the fitness function $f$ subject to the constraint
\begin{displaymath}
  c(X) = |c_S(X) - c_T(X)| \leq B,
\end{displaymath}
where $B$ is an upper bound on how much the number of pixel in $X$ from the two images $S$ and $T$ are allowed to differ.

\section{Evolutionary Algorithm for Image Composition}
\label{sec:ea}

\begin{algorithm}[t]
  \caption{($\mu+1$)~GA for evolutionary image composition}
  \label{alg:eacompo2}
  \begin{algorithmic}[1]
    \Require $S$ and $T$ are images
    \State Initialise population $\mathcal{P} = \{P_1, \ldots, P_{\mu}\}$
    \While{not termination condition}
    \State Select an individual $P_i \in \mathcal{P}$ uniformly at random
    \If{$rand() < p_c$} \Comment{Crossover}
    \State Select $P_{j} \in \mathcal{P} \setminus P_{i}$ uniformly at random
    \If{$rand() < 0.5$}
    \Comment{See Section~\ref{subsec:cross} for $t_{\mathrm{cr}}$}
    \State $Y \gets $ \textsc{RandomWalkMutation($X$,$Z$,$t_{\mathrm{cr}}$)} 
    \Else
    \State $Y \gets $ \textsc{RectangularCrossover($P_i$,$P_j$)}
    \EndIf
    \State $P_i \gets $ \textsc{Selection($P_i$,$Y$)}
    \Else \Comment{Mutation}
    \If{$rand() < 0.5$}
    \State $Y \gets $ \textsc{RandomWalkMutation($P_i$,$S$,$\tmax$)}
    \Else
    \State $Y \gets $ \textsc{RandomWalkMutation($P_i$,$T$,$\tmax$)}
    \EndIf
    \State $P_i \gets $ \textsc{Selection($P_i$,$Y$)}
    \State Adapt \tmax \Comment{See Section \ref{subsec:selfadapt}.}
    \EndIf
    \EndWhile
    \State \textbf{return} $\mathcal{P}$\Comment{Result is a population of evolved images.}
  \end{algorithmic}
\end{algorithm}
  
In this section, we introduce our algorithm for creating artistic images using region covariance descriptors and an evolutionary image transition processes. Our method takes as input two images, $S=(S_{ij})$ and $T=(T_{ij})$ of size $m \times n$, and produces a new images $X=(X_{ij})$ of the same size, that is a mixture of the input images, i.e. $X_{ij} \in \{S_{ij}, T_{ij}\}$ for each $1 \leq i\leq m$ and each $ 1\leq j \leq n$.  The mixed images are generated using the genetic algorithm given in Algorithm~\ref{alg:eacompo2}. The algorithm uses a parent population of $\mu$ images and produces in each iteration one image $Y$ by crossover with probability $p_c$ or by mutation with probability $1-p_c$. The initial population is a multi-set of images of the given images $S$ and $T$. More precisely, each initial individual $P_i$ in the initial population is chosen uniformly at random from $\{S,T\}$.  In order to produce a diverse set of images, an offspring $Y$ is only competing against the first parent $P_i$ for survival. This leads to a low selection process and produces a diverse set of images in the final population. We describe the different components of the genetic algorithm in the following section.

\subsection{Self Adaptive Random Walk Mutation}

Our genetic algorithm uses a variant of the random walk mutation introduced in \cite{neumann17:_evolut}. The mutation operator is shown in Algorithm~\ref{alg:walk}. Given an image $Z$ the random walk starts at a random pixel of the current image $X$ and produces an offspring $Y$ from $X$ by setting each visited pixel $Y_{ij}$ to the value of $Z_{ij}$.

The random walk moves from a current  pixel  $X_{ij}$  either left, right, up, or down to visit the next pixel and does this for \tmax steps.
To formalise this, we define the neighbourhood $N(X_{ij})$ of $X_{ij}$ as
\begin{displaymath}
  N(X_{ij}) = \{ X_{(i-1)j}, X_{(i+1)j}, X_{i(j-1)} X_{i(j+1)} \}
\end{displaymath}
and choose in each step an element of $N(X_{ij})$ uniformly at random.  We use modulo the dimensions of the image which implies that the walk may wrap around the boundaries of the image.

\begin{algorithm}[t]
  \caption{\textsc{RandomWalkMutation($X,Z,\tmax$)}}
  \label{alg:walk}
  \begin{algorithmic}[1]
    \Require $X$ and $Z$ are images
    \State $Y \gets X$
    \State $Y_{ij} \in   Y$ \Comment{Choose starting pixel uniformly at random}
    \State $Y_{ij} \gets Z_{ij}$
    \State $t \gets 1$
    \While{$t \le \tmax$}
    \State Choose $Y_{kl} \in N(Y_{ij})$ uniformly at random
    \State $i \gets k$, $j \gets l$ and $Y_{ij} \gets Z_{ij}$
    \State $t \gets t + 1$
    \EndWhile
    \State \textbf{return} $Y$\Comment{Result is a mutated image.}
  \end{algorithmic}
\end{algorithm}

Given a current image $X$, our genetic algorithm uses the random walk mutation to either paint all the visited pixels with the same values as in $S$ or $T$. Whether to choose $S$ or $T$ is decided uniformly at random for each mutation operation. In this way, the algorithm is able to produce images that are mixtures of $S$ and $T$ and minimise the given fitness function. Each random walk mutation is run for \tmax steps.

\subsubsection*{Self Adaptation}
\label{subsec:selfadapt}

When the algorithm successfully minimises the fitness function, we choose to increase the length of random walks by increasing \tmax. On the other hand, if progress can only be achieved by small random walk mutations, we decrease \tmax.  In particular, we employ the approach for adjusting discrete parameters recently introduced by Doerr and Doerr~\cite{doerr15:_optim}.  Their approach adapts the classical $1/5$-rule adaptation for evolution strategies in \cite{auger09:_bench_es_bbob} to the discrete setting. Our approach increases \tmax in the case of a success and decreases \tmax if the new offspring is not accepted.  With the benefit of a self-adjustable mechanism, \tmax can take on values in ${t_{\mathrm{LB}} \leq t_{\max} \leq t_{\mathrm{UB}}}$, where $t_{\mathrm{LB}}$ is a lower bound on \tmax and $t_{\mathrm{UB}}$ is an upper bound on \tmax.  This differs from the approach in \cite{doerr15:_optim}, where the offspring population size is reduced in the case of a success and increased in the case of a failure.

For a successful mutation, we set
\begin{displaymath}
  t_{\max} \colonequals \min \left\{F\cdot t_{\max}, \, t_{\mathrm{UB}} \right \} 
\end{displaymath}
and for an unsuccessful mutation, we set
\begin{displaymath}
  t_{\max} \colonequals \max \left\{F^{-1/k} \cdot t_{\max}, \, t_{\mathrm{LB}} \right \},
\end{displaymath}
where $F>1$ is a real value and $k\geq 1$ an integer which determines the adaptation scheme.

The scheme implements a $1/(k+1)$-rule where $F$ determines the size of the step changes dependent on the current value of \tmax. For our experimental investigations, we set $t_{\mathrm{LB}}=50$, $t_{\mathrm{UB}}=5000$, $F=2$, $k=8$ based on preliminary experimental investigations. Note that the use of \tmax in Algorithm~\ref{alg:walk} does not require it to be an integer. For a positive real value \tmax, the mutation operator will carry out $\lfloor t_{\max} \rfloor$ steps of the random walk.

\subsection{Crossover}
\label{subsec:cross}

We consider two crossover operators which take two parents and produce one child.

The first crossover operator produces an offspring $Y$ by taking the image of the first parent $P_i$ and performing a random walk for $t_{\mathrm{cr}}$ steps on the image of the second parent $P_j$.  This means that the offspring $Y$ can be produced by calling \textsc{RandomWalkMutation($P_i$, $P_j$, $t_{\mathrm{cr}}$)}.  For our experimental investigations in this paper, we set $t_{\mathrm{cr}}=10000$, i.e.  each random walk crossover mask is generated by a random walk consisting of $10000$ steps.  Note, that this value is independent of \tmax which is adapted during the mutation step. The reason is that we always want to make sure that crossover reduces an offspring which has sufficiently large parts from both parents.

The second crossover operator that we dub \textsc{Rectan\-gu\-larCrossover} produces an offspring where all entries are copied from parent $P_i$ except from a randomly specified rectangle $R$. The upper left point of the rectangle is chosen uniformly at random from the image. The width and the height of the image is chosen as a random integer in $\{1, \ldots, m/10\}$ and $\{1, \ldots, n/10\}$, respectively. Values where the width and/or height would exceed the image boundaries are ignored.

Both crossover operators are focused on a local part of the target image. This facilitates the creation of offspring that have an identified part of the image of the second parent substituted into the image of the first parent. The hope is that the optimisation process benefits from this structured way of creating offspring.

\begin{figure}[!t]
\centering
\begin{tabular}{cccc}
\subcaptionbox*{}{\includegraphics[width = 1.2in]
{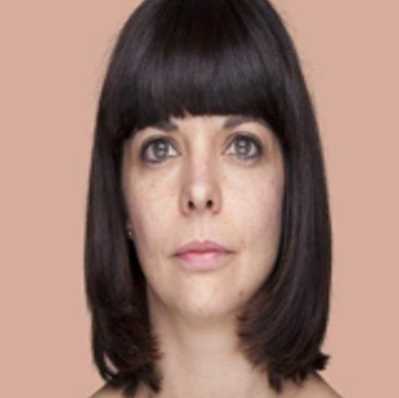}} &
\subcaptionbox*{}{\includegraphics[width = 1.2in]{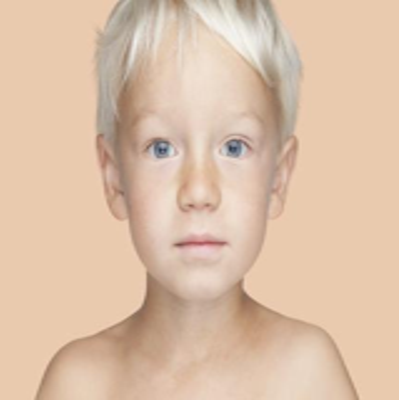}} \\
\subcaptionbox*{}{\includegraphics[width = 1.2in]{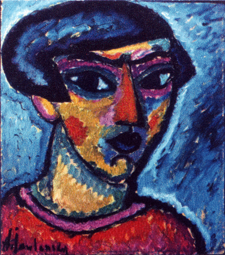}} &
\subcaptionbox*{}{\includegraphics[width = 1.2in]{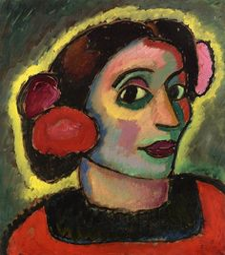}} \\
\subcaptionbox*{}{\includegraphics[width = 1.2in]{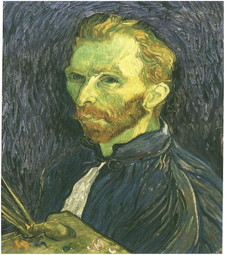}} &
\subcaptionbox*{}{\includegraphics[width = 1.2in]{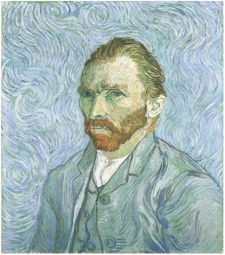}} \\
\end{tabular}
\caption{Pairs of images $S$ (left column) and $T$ (right column).}
\label{source_and_targets}
\end{figure}

\section{Experiments}
\label{sec:exp}

Our experiments were designed to meet a threefold objective: (1) we wished to investigate the impact of different region covariance features on the resulting images; (2) we wanted to discover how different weighting schemes for covariance matrices influence the results; and (3) we wished to understand the influence that the distance measures have on the final results.

\begin{figure*}[!th]
\centering
\includegraphics[width = .2425\textwidth]
{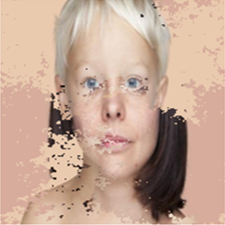}\
\includegraphics[width = .2425\textwidth]
{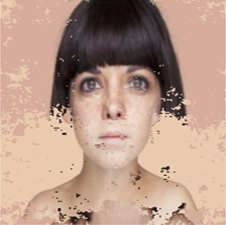}\
\includegraphics[width = .2425\textwidth]
{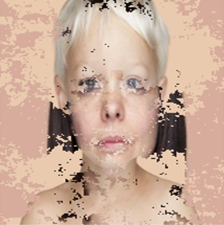}\
\includegraphics[width = .2425\textwidth]
{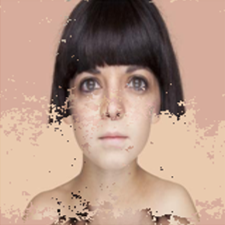}\
\\[0.3ex]
\includegraphics[width = .2425\textwidth]
{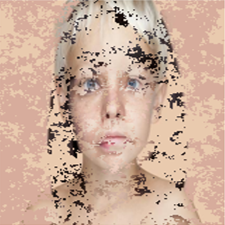}\
\includegraphics[width = .2425\textwidth]
{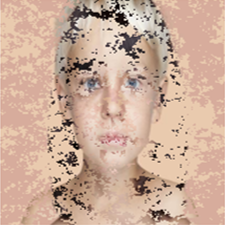}\
\includegraphics[width = .2425\textwidth]
{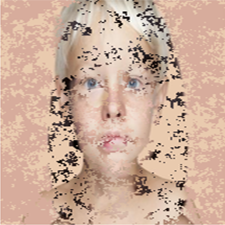}\
\includegraphics[width = .2425\textwidth]
{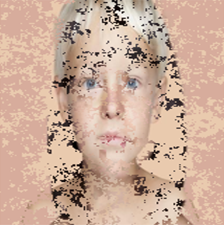}\
\\[0.3ex]
\includegraphics[width = .2425\textwidth]
{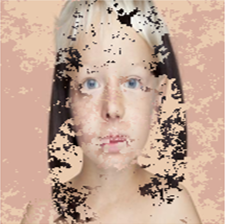}\
\includegraphics[width = .2425\textwidth]
{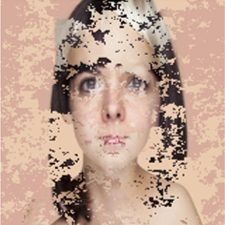}\
\includegraphics[width = .2425\textwidth]
{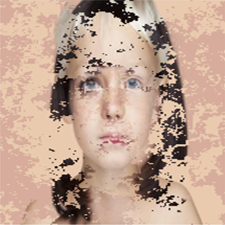}\
\includegraphics[width = .2425\textwidth]
{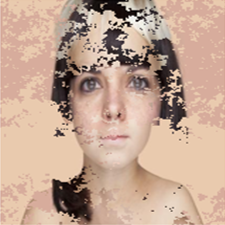}
\caption{Image composition with different features. Rows 1, 2 and 3 correspond to Feature Sets 1, 2 and 3, respectively. Note the structure that emerges with the first feature set.}
\label{faces_features_results}
\end{figure*}

The pairs of images $S$ and $T$ for all of our experiments are depicted in Figure~\ref{source_and_targets}. 
For each setting investigated in this section, we ran our genetic algorithm for $2000$ generations with a population size of $\mu = 4$ and and crossover probability $p_c=0.2$.

\subsection{Impact of Features}

Our first set of experiments were focused on characterising how the choice of features may influence the resulting image. With reference to Table~\ref{tab:1} we chose 3 sets of features as follows:  
\begin{description}
\item[Set 1:] $\left[i,j,r,g,b,\sqrt{\big(\tfrac{\partial I}{\partial i}\big)^2 + \big(\tfrac{\partial I}{\partial j}\big)^2}, \tan^{-1}\left({\big|\frac{\partial I}{\partial i}\big|}\big/{\big|\frac{\partial I}{\partial j}\big|}\right) \right]^\T$ ;
\item[Set 2:] $\left[i,j,h,s,v\right]^\T$;
\item[Set 3:] $\left[h,s,v, \sqrt{\big(\frac{\partial I}{\partial i}\big)^2 + \big(\frac{\partial I}{\partial j}\big)^2}, \tan^{-1}\left({\big|\frac{\partial I}{\partial i}\big|}\big/{\big|\frac{\partial I}{\partial j}\big|}\right) \right]^\T$.
\end{description}  

For all experiments in this section we considered a grid of equispaced square region covariance descriptors ($l = 25$ pixels). We set both sets of weights $w_{(c,d)}^{S}$ and $w_{(c,d)}^{T}$ to $0.5$, and utilised the Euclidean distance as our measure of similarity between covariance matrices. 

The results in Figure~\ref{faces_features_results} demonstrate that the three populations evolve differently depending on the feature set.  Feature Set 1 produced the most visually pleasing results, with the composite image incorporating facial features from both subjects. We attribute this to the correlations that are captured between pixel locations and edge features (magnitude and orientation) in the first feature set, and not in the other two feature sets.

\subsection{Impact of Different Weightings}

\begin{figure}[!t]
\centering
\begin{tabular}{cccc}
\subcaptionbox*{}{\includegraphics[width = 1.5in]{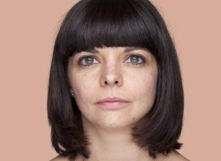}} &
\subcaptionbox*{}{\includegraphics[width = 1.5in]{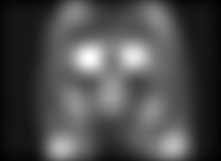}} \\
\end{tabular}
\caption{An image and its corresponding saliency mask.}
\label{s_and_saliency}
\end{figure}

\begin{figure*}[!th]
\centering
\includegraphics[width = .2425\textwidth]
{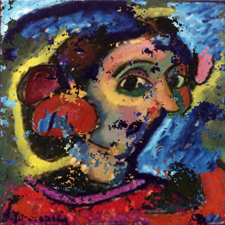}\
\includegraphics[width = .2425\textwidth]
{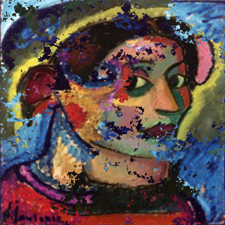}\
\includegraphics[width = .2425\textwidth]
{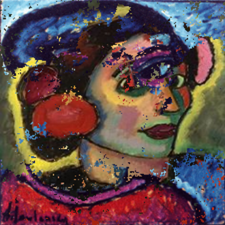}\
\includegraphics[width = .2425\textwidth]
{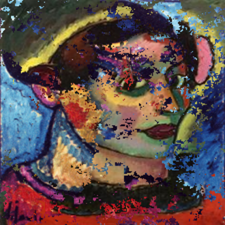}\
\\[0.3ex]
\includegraphics[width = .2425\textwidth]
{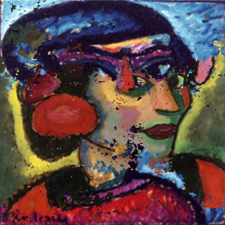}\
\includegraphics[width = .2425\textwidth]
{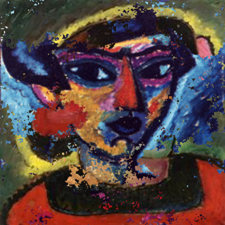}\
\includegraphics[width = .2425\textwidth]
{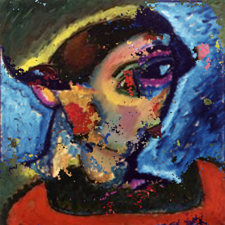}\
\includegraphics[width = .2425\textwidth]
{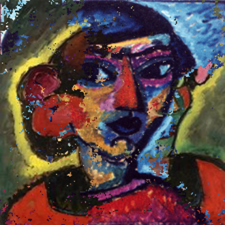}\
\\[0.3ex]
\includegraphics[width = .2425\textwidth]
{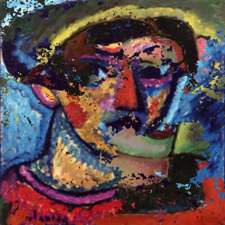}\
\includegraphics[width = .2425\textwidth]
{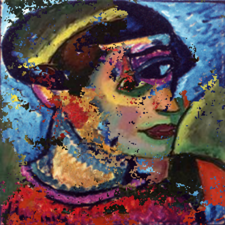}\
\includegraphics[width = .2425\textwidth]
{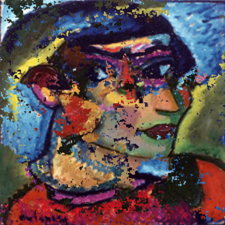}\
\includegraphics[width = .2425\textwidth]
{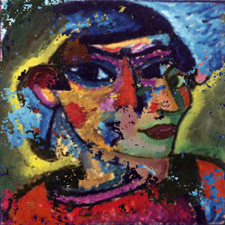}
\\[0.3ex]
\includegraphics[width = .2425\textwidth]
{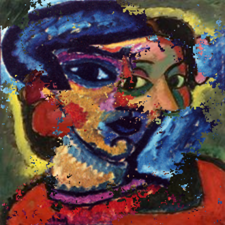}\
\includegraphics[width = .2425\textwidth]
{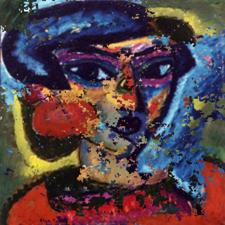}\
\includegraphics[width = .2425\textwidth]
{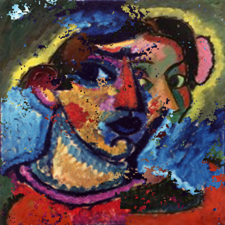}\
\includegraphics[width = .2425\textwidth]
{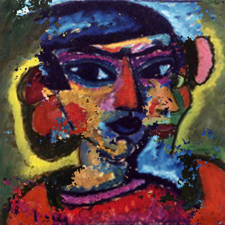}
\caption{Image composition with different covariance weighting schemes. Rows 1, 2 and 3 correspond to $w_{(c,d)}^{S}$ set to $0.25$, $0.5$ and $0.75$ and  $w_{(c,d)}^{T}$ set to $0.75$, $0.5$ and $0.25$, respectively. In the last row the weights were set using an image saliency algorithm. The saliency algorithm strikes a consistent balance between notable regions in both images.}
\label{weight_experiments}
\end{figure*}

In the second set of experiments we explored the consequence of using different weights for the region covariance matrices. We limited ourselves to two weighting schemes: uniform and saliency-based. In the uniform regime all of the weights $w_{(c,d)}^{S}$ associate with $S$  were set to the same value, and, similarly, all the weights $w_{(c,d)}^{T}$ associated with  $T$  were also fixed. In particular,  for simulations 1, 2 and 3, we set $w_{(c,d)}^{S}$  to $0.2$, $0.5$ and $0.75$ and $w_{(c,d)}^{T}$ to $0.75$, $0.5$ and $0.25$, respectively. 

In contrast, for a simulation with saliency-based weighting, we utilised the image saliency algorithm of \cite{hou2012image} to assign weights for each region covariance descriptor. The purpose of the saliency algorithm is to predict human fixation points, which in turn are used as a measure of visual attention. The saliency algorithm takes as input an image, and produces a new image of the same size which can be interpreted as a probability map designating which regions of an image a human would pay most of their attention to (see Figure~\ref{s_and_saliency}). We ran the saliency algorithm on both $S$ and $T$. The weights for $w_{(c,d)}^{S}$ and $w_{(c,d)}^{T}$  were taken from the saliency maps associated with $S$ and $T$, respectively. 

For all experiments in this section we used Feature Set 1 and employed a grid of equispaced square region covariance descriptors ($l = 20$ pixels). We measured the similarity between covariance matrices using the Log-Euclidean distance.

The results of our experiment are depicted in Figure~\ref{weight_experiments}. As expected, when uniform weights are used the global relative contribution of the image $S$ can be traded-off against the global relative contribution of the image $T$. The benefit of the saliency-based weighting scheme is that it can automatically produce visually pleasing results that strike a balance between notable regions in both images. 

\subsection{Impact of Distance Measures}

\begin{figure*}[!t]
\centering
\includegraphics[width = .2425\textwidth]
{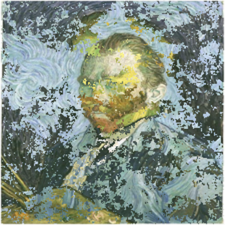}\
\includegraphics[width = .2425\textwidth]
{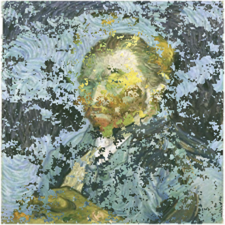}\
\includegraphics[width = .2425\textwidth]
{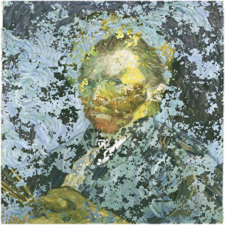}\
\includegraphics[width = .2425\textwidth]
{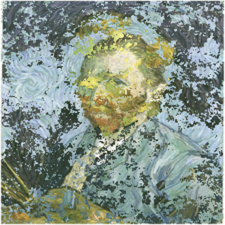}\
\\[0.3ex]
\includegraphics[width = .2425\textwidth]
{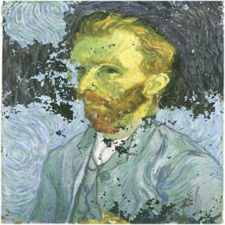}\
\includegraphics[width = .2425\textwidth]
{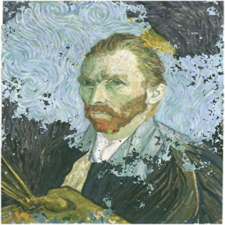}\
\includegraphics[width = .2425\textwidth]
{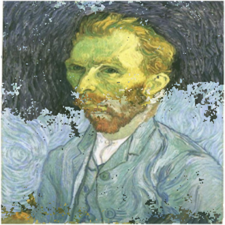}\
\includegraphics[width = .2425\textwidth]
{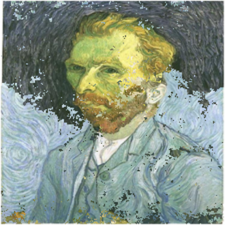}\
\\[0.3ex]
\includegraphics[width = .2425\textwidth]
{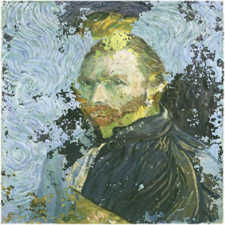}\
\includegraphics[width = .2425\textwidth]
{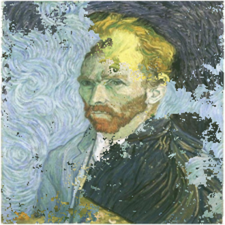}\
\includegraphics[width = .2425\textwidth]
{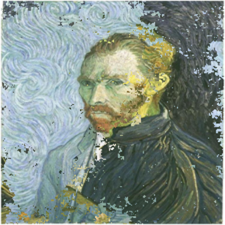}\
\includegraphics[width = .2425\textwidth]
{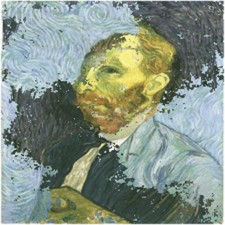}
\caption{Image composition with different covariance distances. Rows 1, 2 and 3 correspond to distance metrics $\dist{E}$, $\dist{A}$ and $\dist{L}$, respectively. Note how the Euclidean distance yields patchy results, whereas the other distance measures result in more structured images. }
\label{distance_experiments}
\end{figure*}

In the third and final set of experiments we investigated the utility of different metrics for covariance matrices. Once again, a notable structure emerges. The Euclidean distance measure produced inferior results, with the images taking on a very patchy and noisy appearance. In contrast, the images generated using the Log-Euclidean or affine-invariant distance were much more cohesive and interesting. The results presented in Figure~\ref{distance_experiments} are based on grid of equispaced square region covariance descriptors ($l = 20$ pixels) using saliency-based weights and Feature Set 1. 

\section{Discussion}

\begin{figure*}[!th]
\centering
\includegraphics[width = .2425\textwidth]
{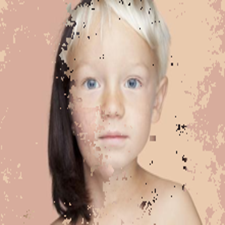}\
\includegraphics[width = .2425\textwidth]
{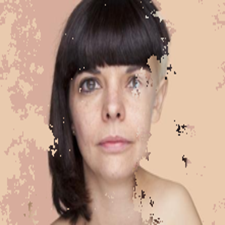}\
\includegraphics[width = .2425\textwidth]
{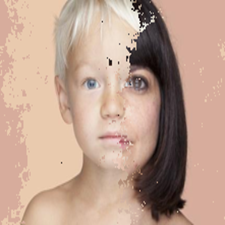}\
\includegraphics[width = .2425\textwidth]
{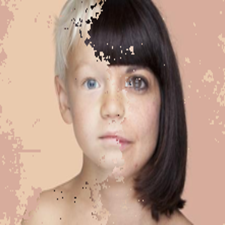}\
\\[0.3ex]
\includegraphics[width = .2425\textwidth]
{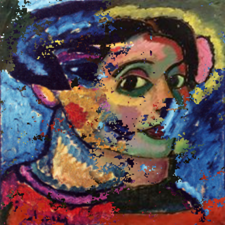}\
\includegraphics[width = .2425\textwidth]
{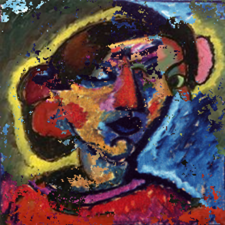}\
\includegraphics[width = .2425\textwidth]
{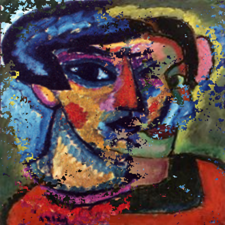}\
\includegraphics[width = .2425\textwidth]
{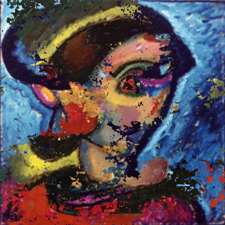}\
\\[0.3ex]
\includegraphics[width = .2425\textwidth]
{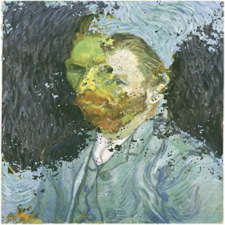}\
\includegraphics[width = .2425\textwidth]
{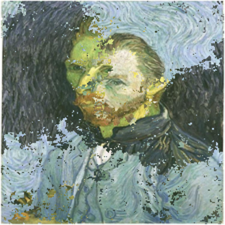}\
\includegraphics[width = .2425\textwidth]
{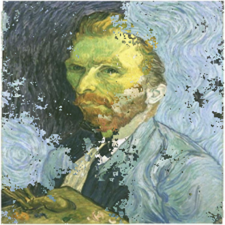}\
\includegraphics[width = .2425\textwidth]
{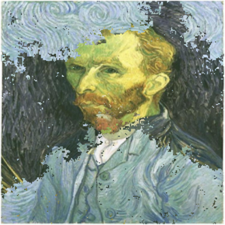}
\caption{Image composition using Feature Set 1 with saliency-based weighting and the Log-Euclidean distance measure.}
\label{combo_results}
\end{figure*}

Our experimental examinations suggest that the best results are obtained when Feature Set 1 is used in conjunction with saliency-based weighting and a Log-Euclidean distance measure. To verify this hypothesis, we deployed our genetic algorithm one more time with these parameters on all three image pairs. The results presented in Figure~\ref{combo_results} confirm our hypothesis. In all cases, our algorithm produced a population of new aesthetic images that incorporate pertinent regions from both input images.  

\section{Conclusion}

We have introduced a new approach of image composition based on feature covariance matrices. This approach facilitates the composition of novel artistic images. We have introduced a genetic algorithm evolving a set of images by crossover and mutation operators based on random walk. For our experimental investigations, we have considered different pairs of images and compared the final populations showing the resulting images with respect to different parameters.

\section*{Acknowledgements}

This work has been supported through Australian Research Council (ARC) grants DP140103400 and LE160100090.

\bibliographystyle{ACM-Reference-Format}
\bibliography{gecco2017-acm-bib}

\end{document}